\documentclass[letterpaper]{article} 
\usepackage{aaai24}  
\usepackage{times}  
\usepackage{helvet}  
\usepackage{courier}  
\usepackage[hyphens]{url}  
\usepackage{graphicx} 
\usepackage{natbib}  
\usepackage{caption} 
\usepackage{algorithm}

\usepackage{amsmath}
\usepackage{amsfonts}
\usepackage{algorithm}
\usepackage{algpseudocode}
\usepackage{comment}
\usepackage{booktabs}
\usepackage{multirow}

\newcommand{\norm}[1] { \left\Vert #1 \right\Vert }

\newtheorem{proposition}{Proposition}
\newtheorem{definition}{Definition}
\usepackage{newfloat}
\usepackage{listings}
\DeclareCaptionStyle{ruled}{labelfont=normalfont,labelsep=colon,strut=off} 
\floatstyle{ruled}
\newfloat{listing}{tb}{lst}{}
\floatname{listing}{Listing}
\title{Online Sensitivity Optimization in Differentially Private Learning}
\author {
    Filippo Galli,\textsuperscript{\rm 1,4}
    Catuscia Palamidessi,\textsuperscript{\rm 2,3}
    Tommaso Cucinotta\textsuperscript{\rm 4}
}
\affiliations {
    \textsuperscript{\rm 1}Scuola Normale Superiore\\
    \textsuperscript{\rm 2}INRIA, Palaiseau, France\\
    \textsuperscript{\rm 3}\'Ecole Polytechnique, Palaiseau, France\\
    \textsuperscript{\rm 4}Scuola Superiore Sant’Anna, Pisa, Italy\\
    filippo.galli@sns.it, catuscia@lix.polytechnique.fr, tommaso.cucinotta@santannapisa.it
}

\nocopyright
\begin{document}

\maketitle

\begin{abstract}
Training differentially private machine learning models requires constraining
an individual's contribution to the optimization process. 
This is achieved by clipping the $2$-norm of their gradient at a predetermined
threshold prior to averaging and batch sanitization.
This selection adversely influences optimization in two opposing ways: 
it either exacerbates the bias due to excessive clipping at lower values,
or augments sanitization noise at higher values. The choice significantly
hinges on factors such as the dataset, model architecture, and even varies
within the same optimization, demanding meticulous tuning usually
accomplished through a grid search.
In order to circumvent the privacy expenses incurred in hyperparameter 
tuning, we present a novel approach to dynamically optimize the clipping
threshold. We treat this threshold as an additional learnable parameter, 
establishing a clean relationship between the threshold and the cost
function. This allows us to optimize the former with gradient descent, with
minimal repercussions on the overall privacy analysis.
Our method is thoroughly assessed against alternative fixed and adaptive strategies
across diverse datasets, tasks, model dimensions, and privacy levels.
Our results indicate that it performs comparably or better in the 
evaluated scenarios, given the same privacy requirements.

\end{abstract}

\section{Introduction} \label{sec:intro}
The widespread adoption of machine learning techniques has led to increased concerns about user
privacy. Users who share their data with service providers are becoming more cautious due to 
the exposure of various privacy attacks, observed in both academic and industrial contexts
\cite{carlini2021extracting, carlini2023extracting, papernot2018sok}.
As a result, there is a growing effort to enhance training methods that can provide strong 
and quantifiable privacy guarantees.

In the privacy-preserving machine learning community, differential privacy \cite{dwork2006differential}
has emerged as the predominant framework for defining privacy requirements and strategies.
Essentially, training a model with differential privacy requires bounding the contribution of 
a single individual to the overall procedure, to guarantee that the trained model will be 
probabilistically indistinguishable compared to the same model trained without including 
any one specific user in the dataset. In the context of gradient-based learning,
this is achieved by introducing a parameter $C$ called the \textit{clipping threshold} \cite{abadi2016deep}.
This parameter controls the magnitude of gradients from each user (or sample) before 
they are averaged with contributions from other users. Subsequently, the result
undergoes a process of differential privacy sanitization, which involves the addition 
of random Gaussian noise proportional to the value of $C$.

\begin{figure}[t]
    \centering
    \includegraphics[scale=0.4]{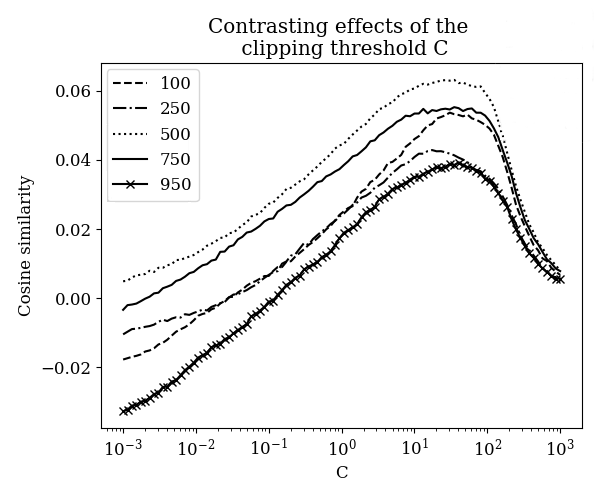}
    \caption{The choice of clipping threshold $C$ requires trading off a higher 
    clipping bias at small values, for larger Gaussian noise at large values. 
    Here the clipped, averaged, noised gradient of a CNN for
    character recognition is compared with the true average gradient at 
    different training iterations $t \in \{100, 250, 500, 750, 950\}$. Note that
    for some values the sanitized gradient may even have components pointing in the 
    opposite direction w.r.t the true gradient, corresponding to negative
    cosine similarity. The reported figure of cosine similarity is an 
    average over $20$ realizations of the Gaussian mechanism.}
    \label{fig:trade-off}
\end{figure}

The choice of the clipping threshold $C$ is crucial: on the one hand, large values introduce noise
levels that may slow down or hinder the optimization altogether; on the other hand, 
small values introduce a bias in the average clipped gradient with respect to the true average gradient
and may leave the optimization stuck in bad local minima. Figure \ref{fig:trade-off} exemplifies 
the issue.
Note that the clipping bias is not only directed toward zero (as bounding the $2$-norm may
lead to believe), but depends, in general, on the distribution of
the per-sample gradients around the expectation \cite{chen2020understanding}.
Achieving an optimal trade-off remains an ongoing challenge.
Historically, researchers have treated 
the clipping threshold as a parameter to be optimized, often through a grid or random search, 
in order to assess the performance of privacy preserving models in the ideal conditions 
in which an oracle provides the optimal values for the hyperparameters. 
However, it is worth noting that every additional gradient-query to the dataset for optimization
purposes introduces a certain degree of privacy leakage.
Of late though, the implications of not accounting for privacy leakage over
multiple runs of a grid search have drawn more attention, leading to different accounting
strategies \cite{papernot2022hyperparameter, mohapatra2022role, liu2019private}. 
The inherent challenges of increased privacy leakage and computational overhead resulting
from extensive hyperparameter searches persist, necessitating further innovations to 
encourage broader adoption of differentially private machine learning techniques.
Therefore, we set out to find a strategy for the online optimization
of the clipping threshold that is privacy preserving and computationally inexpensive,
while maintaining comparable or better performance on a set of tasks, datasets, 
and model architectures. 

\textbf{Our contributions}: i) We investigate the sensitivity trade-off in differentially
private learning in terms of cosine similarity between the sanitized and true gradients,
showing that at every iteration it is possible to determine a fairly prominent optimal
value, ii) we elaborate a strategy for the online optimization of the sensitivity, 
taking from the literature in online learning rate optimization and extending it to
optimize the clipping threshold, iii) we establish the corresponding techniques for doing
so privately, which require allocating a marginal privacy budget and iii) we provide experimental results
to validate our algorithm in multiple contexts and against a number of relevant state-of-the-art strategies for private hyperparameter optimization.

\section{Background} 
\label{sec:back}
Gradient-based optimization of supervised machine learning models typically implies 
finding the optimal set of parameters $\theta \in \mathbb{R}^n$ to fit a function 
$f_{\theta}: \mathcal{X} \rightarrow \mathcal{Y}$ to a dataset $D \in \mathcal{D}$
of pairs $z_i = (x_i,y_i) \in \mathcal{X}\times \mathcal{Y}$, 
by minimizing an error function $\ell: \mathbb{R}^n \times \mathcal{D} 
\rightarrow \mathbb{R}_{\geq 0}$. At time $t$, the iterative optimization process
computes the cost of mismatched predictions and updates the parameters towards
the nearest local minimum of $\ell$ by repeated applications of the (stochastic) 
gradient descent algorithm $\theta_{t+1} \leftarrow \theta_t - \rho g_t$, with $\rho$ 
the learning rate, and 
\begin{equation}
\label{eq:avg_grad}
g_t = \frac{1}{|B|}\sum_{z_i \in B}\nabla_{\theta_t}
\ell(f_{\theta_t}(x_i), y_i)
\end{equation}
being the average gradient of the 
error function with respect to the parameters, computed over
the samples $ z_i = (x_i, y_i)$ of the minibatch $B \subseteq D$.

As is now commonplace in the machine learning literature, 
privacy guarantees are provided within the 
framework of Differential Privacy (DP):
\begin{definition}[Differential Privacy \cite{dwork2006differential}]
\label{def:dp}
A randomized mechanism $\mathcal{M}: \mathcal{D} \rightarrow \mathcal{R}$
with domain $\mathcal{D}$ and range $\mathcal{R}$ satisfies $(\varepsilon, \delta)$
differential privacy if for any two datasets $D, D' \in \mathcal{D}$ differing in at most one sample,
and for any outputs $S \subseteq \mathcal{R}$ it holds that
\begin{equation}
    \text{Pr}[\mathcal{M}(D) \in S] \leq e^{\varepsilon} \text{Pr}[\mathcal{M}(D') \in S] + \delta
\end{equation}
\end{definition}
To fit machine learning optimization within the definition of a differentially
private random mechanism (intended here in a broad sense to also 
include later generalizations \cite{dwork2016concentrated, mironov2017renyi})
the average gradient in Equation 
\eqref{eq:avg_grad} is sanitized by means of the Gaussian mechanism 
\cite{dwork2014algorithmic}. 
In particular, if $h: \mathcal{X} \rightarrow \mathbb{R}^n$ is a function
with $2$-norm sensitivity $S_h$, the DP approximation $\tilde{h}(x)$ of
$h(x)$, $x\in \mathcal{X}$, can be found as 
\begin{equation}
\label{eq:gm}
    \tilde{h}(x) = h(x) + \eta, \quad \eta \sim \mathcal{N}(0, \sigma^2I=S_h^2\nu^2I)
\end{equation}
with $\nu$ 
the noise multiplier which depends only on the privacy parameters, and 
$\mathcal{N}$ being a random normal distribution. Tuning the 
additive Gaussian noise implies tuning its standard deviation proportionally to 
the $2$-norm sensitivity of the query $g_t$ over the minibatch $B$.
As, in general,  $\norm{g_t}_2$ is not bounded \textit{a priori}, the per-sample 
gradients of the error function are clipped in norm to a 
certain value $C_t$ \cite{song2013stochastic, bassily2014private, shokri2015privacy,
abadi2016deep} by applying the transformation 
\begin{equation}
    \bar{g}_t(z_i) = \frac{g_t(z_i)}{
    \text{max}\left(1, \frac{\norm{g_t(z_i)}_2}{C_t}\right)}
\end{equation}
from which follows the sensitivity of the average clipped gradient, allowing 
for the sanitization of the query at the $t^{\text{th}}$ iteration. 
The Gaussian mechanism lends itself to a refined analysis of the privacy 
leakage incurred in its repeated application, which is essential in practical 
machine learning with stochastic gradient descent to keep the overall privacy
expenditure to a minimum  over multiple training epochs 
\cite{abadi2016deep, wang2019subsampled}.
A similar procedure can be utilized to account for multiple runs with different
configurations in a grid search \cite{mohapatra2022role}.

\section{Related Works} \label{sec:rel}
This work draws from two main lines of research, namely hyperparameter
optimization in non-private settings and sensitivity optimization 
in differentially private machine learning.
\subsubsection{Hyperparameter Optimization}
Sub-gradient minimization strategies such as SGD iteratively approach the optimal 
solution by taking steps in the direction of steepest descent of a cost function. 
For this heuristic to be effective, the length of each step needs to be tuned by 
controlling the \textit{learning rate}, which has been considered the ``single most
important hyperparameter" \cite{bengio2012practical}. Many works have introduced 
strategies for its adaptive tuning, such as \cite{lydia2019adagrad, 
kingma2014adam}, which adjust the per-parameter value w.r.t. a common value still
defined \textit{a priori}. Conversely, other research has exploited automatic
differentiation to concurrently optimize the parameters and hyperparameters
\cite{maclaurin2015gradient} via SGD. In particular, explicitly deriving the 
partial derivative of the cost function with respect to the learning rate has 
been demonstrated to be an effective strategy, and it has been discovered
independently at different times \cite{almeida1999parameter, gunes2018online}.
These works do not explore the private setting and introduce general methods
that are almost exclusively applied to learning rate optimization, without
addressing the choice of other hyperparameters. In \cite{mohapatra2022role}
instead, the authors study adaptive optimizers in the differentially private setting,
by analyzing the estimate of the raw second moment of the gradient at convergence.
Their objective is to reduce the privacy cost of tuning the learning rate in a 
grid search, but the clipping threshold is still treated as an additional 
hyper-parameter.

\subsubsection{Sensitivity Optimization}
As discussed in the Introduction and Background sections,
establishing the value of the clipping 
threshold $C_t$ is 
critical in differentially private machine learning, and treating this value as a 
hyper-parameter has largely been the preferred strategy in the literature 
\cite{song2013stochastic, bassily2014private, shokri2015privacy, abadi2016deep}.
Grid searching over the candidate values can be tricky as gradient norms may span
many orders of magnitude and the effects of more aggressive clipping are not 
easily predicted before running an optimization. Considering also the increased 
privacy costs of running multiple configurations, hyperparameter
selection under privacy constraints is a thriving research area 
\cite{papernot2022hyperparameter, liu2019private, mohapatra2022role}.

Adaptive clipping strategies have also been considered. \cite{andrew2021differentially}
updates $C_t$ during training to match a target quantile of the gradient norms,
which is fixed beforehand.
Although the optimal quantile is still a hyper-parameter, 
its domain is limited to the $[0, 1] \subset \mathbb{R}$ interval. 
Moreover, \cite{andrew2021differentially} shows that adaptively
updating $C_t$ outperforms even the best fixed-clipping strategy.
Additionally, as DP training has shown to disproportionally favor majority classes 
in a dataset \cite{suriyakumar2021chasing}, tuning a target quantile instead of a fixed
clipping threshold may help at least in quantifying the issue, if not in solving it.
Note that although this strategy was introduced to train differentially
private \textit{federated} machine learning
models, the attacker is still modelled as an \textit{honest-but-curious}
adversary and thus it relies on a central trusted server to provide DP guarantees. 
Therefore, the clipping strategy in \cite{andrew2021differentially}
can be used to train \textit{centralized} 
machine learning models just by switching from user-level to sample-level
differential privacy \cite{mcmahan2018learning}.
To further stress this point, note that 
although in \cite{andrew2021differentially} each single user clips the update and sends 
statistics to the central 
server, from a differential privacy point of view this is identical to the server 
performing these operations itself on the true per-user gradients.

\section{Method} \label{sec:meth}
Inspired by the literature on online hyperparameter optimization discussed
in the Related Works,
the idea behind this method is to optimize the clipping threshold based
on the chain rule for derivatives, so that we can find what change in $C_t$ will 
induce a decrease in the cost function $\ell(\theta_t)$. 
Although this strategy works in general for sub-gradient
methods, we are going to explicitly derive the results for DP-SGD. 
Given the SGD update rule with gradient clipping:
\begin{align} \label{eq:sgd}
    \theta_{t+1} & = \theta_t - \rho \nabla \ell(\theta_t) \\
    & = \theta_t - \rho \frac{1}{|B_t|} \sum_{z_i \in B_t} 
    \frac{g_t(z_i)}{\max (1, \norm{g_t(z_i)}_2/C_t)}
\end{align}
we want to find
\begin{align}
    \frac{\partial \ell(\theta_t)}{\partial C} & = 
    \frac{\partial \ell(\theta_t)}{\partial \theta_t}^\top\frac{\partial \theta_t}{\partial C_t} \\
    & = \nabla \ell(\theta_t)^\top\frac{\partial \theta_t}{\partial C_t} \\
    & = -\rho\nabla \ell(\theta_t)^\top\frac{\partial \nabla \ell(\theta_{t-1})}{\partial C_{t-1}} \label{eq:dfdc}
\end{align}
where in the last equality we exploit $\theta_t = \theta_{t-1} - \rho \nabla \ell(\theta_{t-1})$ and assume
$C_t \approx C_{t-1}$. To find an explicit form for Equation \eqref{eq:dfdc}, we notice that the rightmost
term is differentiable almost everywhere, with:
\begin{equation}
    \frac{\partial \nabla \ell(\theta_{t-1})}{\partial C_{t-1}} = q_{t-1} = 
    \frac{1}{|B_{t-1}|} \sum_{z_i \in B_{t-1}} q_{t-1}(z_i)
\end{equation}
and
\begin{equation}
    q_{t-1}(z_i) =
    \begin{cases}
        \frac{g_{t-1}(z_i)}{\norm{g_{t-1}(z_i)}_2} & \text{if} \norm{g_{t-1}(z_i)}_2 > C_{t-1} \\
        0 & \text{otherwise}
    \end{cases} \label{eq:q}
\end{equation}
where we highlight that $\norm{q_{t-1}(z_i)}_2 \in \{0, 1\}, \forall z_i \in B_{t-1}$ by definition of the
clipping function.
Thus we find:
\begin{equation}
    \frac{\partial \ell(\theta_t)}{\partial C_t} = -\rho\nabla \ell(\theta_t)^\top
    \frac{1}{|B_{t-1}|} \sum_{z_i \in B_{t-1}} q_{t-1}(z_i) \label{eq:dfdc-explicit}
\end{equation}
resulting in the gradient descent update rule for the clipping threshold:
\begin{equation}
    C_{t+1} = C_t + \rho_c\rho\nabla \ell(\theta_t)^\top
    \frac{1}{|B_{t-1}|} \sum_{z_i \in B_{t-1}} q_{t-1}(z_i) \label{eq:additive}
\end{equation}
which is the dot product of the current average gradient with a masked
version of last iteration's average gradient, where all per-sample
gradients have either norm $0$ or $1$.

Taking into account the coupled dynamics of the learning rate and the clipping threshold
\cite{mohapatra2022role}, having an adaptive clipping strategy may still slow down 
convergence if the learning rate is kept fixed at the starting value. Thus, we use the 
same method to derive an update strategy for the learning rate $\rho_t$, as in 
\cite{almeida1999parameter, gunes2018online}:
\begin{align}
    \frac{\partial \ell(\theta_t)}{\partial \rho_t} & = 
    \frac{\partial \ell(\theta_t)}{\partial \theta_t}^\top\frac{\partial \theta_t}{\partial \rho_t} \\ 
    & = \nabla \ell(\theta_t)^\top \nabla \ell(\theta_{t-1})
\end{align}
which results in the dot product of the current and past clipped gradients, yielding:
\begin{equation}
\label{eq:lrlr}
    \rho_{t+1} = \rho_t + \rho_r\rho\nabla \ell(\theta_t)^\top \nabla \ell(\theta_{t-1})
\end{equation}

We do not
further expand this result for brevity and because computing these quantities does
not require a dedicated procedure, as they are already a byproduct 
of SGD to optimize $\theta_t$, even in a non-private setting.

\section{Privacy Analysis} \label{sec:priv}
When assuming a time-dependent $C_t$ such as in 
\cite{andrew2021differentially}, it is particularly useful to decouple the contributions of 
the sensitivity from contributions of the privacy parameters $(\varepsilon, \delta)$ 
to the variance of the Gaussian mechanism, as in Equation \eqref{eq:gm}.
Then, within the framework of Rényi DP and given the results in
\cite{mironov2017renyi, wang2019subsampled} one can efficiently determine 
ahead of training-time the values of noise multiplier to be applied at
each iteration independently of the current value of $C_t$.
At the $t^{\text{th}}$ iteration there may be two sources of differential privacy leakage: 
the computation of $\theta_{t+1}$ in Equation \eqref{eq:sgd} and the computation of $C_{t+1}$
in Equation \eqref{eq:additive}. Both can be sanitized with the
DP approximation already discussed, but the latter needs special attention.
To sanitize $C_{t+1}$ with the Gaussian mechanism 
(for reasons detailed in Proposition \ref{prop}) we may utilize 
$\nabla \ell(\theta_t) \approx \nabla \tilde{\ell}(\theta_t)$, effectively repurposing
the sanitized gradient with respect to $\theta_t$. We focus now on
the non-privatized term $\partial \nabla \ell(\theta_{t-1})/\partial C_{t-1}$.
Naturally, it still involves the sanitization of a sum of vectors, 
with the fortunate benefit of having all the terms in the summation be of norm 
either $0$ or $1$, as shown in Equation \eqref{eq:q}, resulting in the unit 
sensitivity of the query. Thus, this step does not introduce the need to develop
any further ``higher order" (adaptive) clipping strategies.
With the considerations above, from a privacy perspective, the two privatized
parallel queries behave as a single query sanitized with the Gaussian mechanism.
This result is formalized in Proposition \ref{prop}, which follows from the 
\textit{joint clipping} strategy described in \cite{mcmahan2018general}.

\begin{proposition}
\label{prop}
The Gaussian approximations $\tilde{q}_t$ and $\tilde{g}_t$ of
$\sum_{z_i \in B_{t-1}} q_{t-1}(z_i)$ and $\sum_{z_i \in B_t} \bar{g}_t(z_i)$
with noise multipliers, respectively, $\nu_q$ and $\nu_g$, is equivalent (as far as privacy accounting is
concerned) to the application of a single
Gaussian mechanism with noise multiplier $\nu$ if $\nu_g = (\nu^{-2} - \nu_q^{-2})^{-1/2}$.
\end{proposition}

\begin{figure}
    \centering
    \includegraphics[scale=0.172]{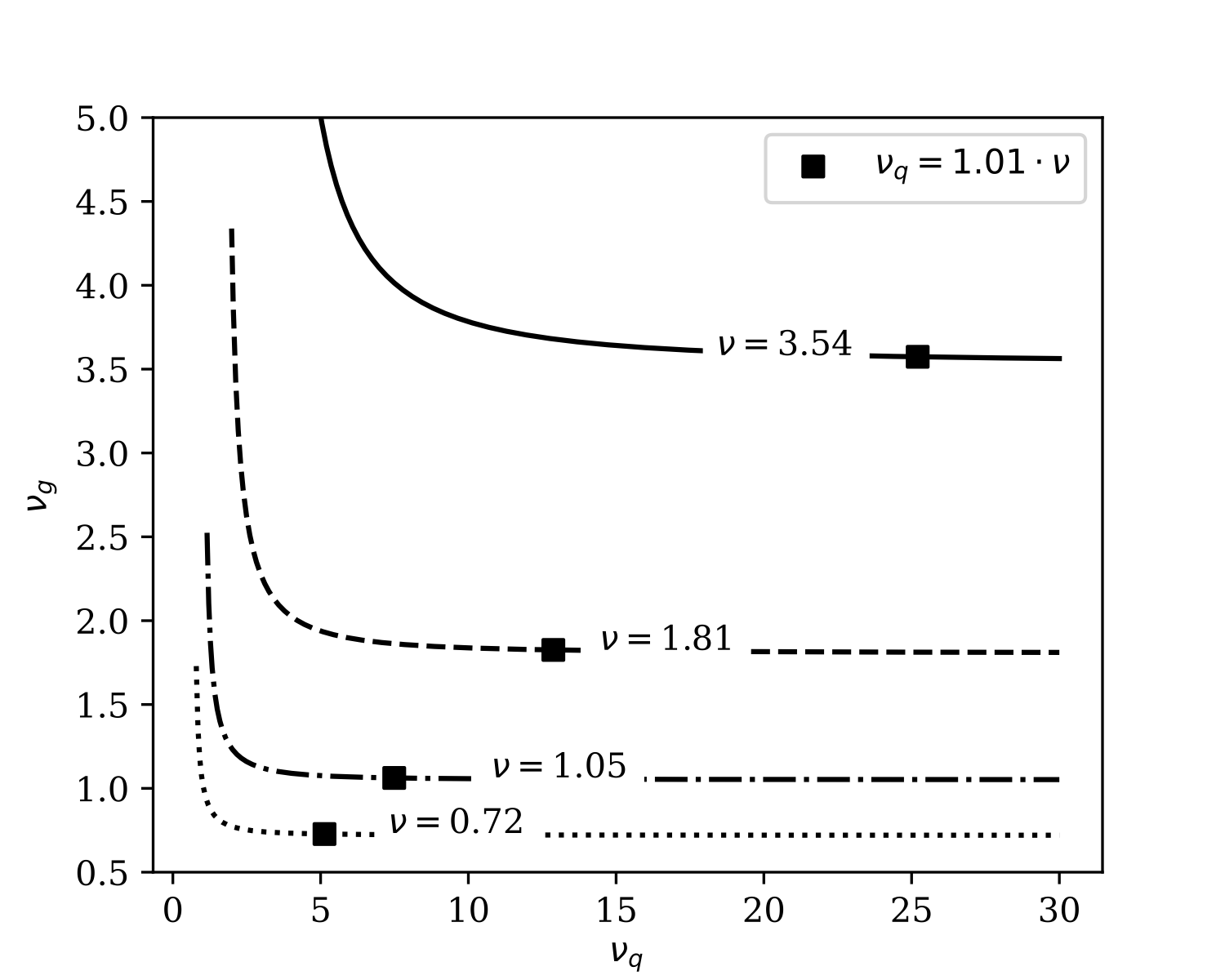}
    \caption{The Pareto frontiers of the noise multipliers to sanitize $\tilde{g}_t$
    and $\tilde{q}_t$, and the chosen values given the heuristic
    described in the Privacy Analysis section, at different privacy requirements.
    This particular instance comes from the MNIST experiments described
    in the Experiments section.}
    \label{fig:nu-trade-off}
\end{figure}

Compared to Theorem 1 in \cite{andrew2021differentially}, we lose a factor of $2$ in the
reduction of the standard deviation $\sigma_q = 1\cdot \nu_q$ and 
since $\nu_q$ is used here to 
sanitize a sum of vectors in $\mathbb{R}^n$ (whereas 
\cite{andrew2021differentially} only need to sanitize a scalar quantity)
we cannot relegate as much differentially private noise to the computation of 
$\tilde{q}_t$. Nonetheless, we can derive a rule of thumb, which, together 
with practical considerations introduced in the next section, allow to
have working estimates of the true $\partial \ell(\theta_t)/\partial C_t$.
In particular, if we allow a $1\%$ increase in $\nu_g$ over $\nu$, we can 
rearrange the result in Proposition \ref{prop} to find 
$\nu_q \approx 7.124\cdot \nu$. 
Figure \ref{fig:nu-trade-off} shows an example of these trade-offs
for the MNIST dataset discussed in Experiments section. 

To complete the privacy analysis, we highlight that from a DP point
of view, the updates to the learning rate described in Equation \eqref{eq:lrlr}
come with no additional privacy expenditure with respect to DP-SGD, exploiting the sanitized $\tilde{g}_t$ and $\tilde{g}_{t-1}$.

\section{The OSO-DPSGD Algorithm} \label{sec:algo}
The algorithm keeps track of two sanitized quantities at each iteration, that is:
\begin{equation}
    \tilde{q}_t = \frac{1}{|B_t|} \sum_{z_i \in B_t} q_t(z_i) + \eta, \quad \eta \sim \mathcal{N}(0, \nu_q^2I)
\end{equation}
\begin{equation}
   \tilde{g}_t = \frac{1}{|B_t|} \sum_{z_i \in B_t} \bar{g}_t(z_i) + \eta, \quad \eta \sim  \mathcal{N}(0, C_t^2\nu_g^2I)
\end{equation}
from which one can privately compute the parameter update and
$\partial \tilde{\ell}(\theta_t)/\partial C_t = -\rho\tilde{g}_t^\top
\tilde{q}_{t-1}$, which requires to store $\tilde{q}_{t-1}$ from the last iteration. 
Note that storing vectors from past iterations is a common strategy even in
non-privatized learning, as e.g. it is required by every optimizer with momentum(s).
In order to cater to the wide range of values $C_t$ might take, spanning orders of
magnitude \cite{andrew2021differentially}, instead of relying on the additive update
rule in Equation \eqref{eq:additive}, we first consider the scale-invariant 
Equation \eqref{eq:priv-C} proposed in \cite{rubio2017convergence},
which converges with a logarithmic number of steps, instead of linearly
\begin{equation} \label{eq:priv-C}
    C_{t+1} = C_t\cdot \left(1 + \rho_c \frac{\tilde{g}_{t}{}^\top \tilde{q}_{t-1}}
    {\norm{\tilde{g}_{t}}_2\norm{\tilde{q}_{t-1}}_2} \right)
\end{equation}
We briefly experimented with Equation \eqref{eq:priv-C} and found the 
proportional update step
$\tilde{g}_{t}{}^\top \tilde{q}_{t-1}/\norm{\tilde{g}_{t}{}^\top\tilde{q}_{t-1}}_2 =
\text{sign}(\tilde{g}_{t}{}^\top \tilde{q}_{t-1})$
to be more robust w.r.t. the Gaussian noise and less dependent on the particular
choice of $\rho_c$. Noticing that $1 + x \approx e^x$ for small values of $x$, we converge
to an exponential update rule for the optimization of both $C_t$ and $\rho_t$, similar to \cite{andrew2021differentially}:
\begin{equation}
\label{eq:exp_C}
    C_{t+1} = C_t\cdot \text{exp}(\rho_c\text{sign}(\tilde{g}_{t}{}^\top \tilde{q}_{t-1}))
\end{equation}
\begin{equation}
\label{eq:exp_r}
    \rho_{t+1} = \rho_t\cdot \text{exp}(\rho_r\text{sign}(\tilde{g}_{t}{}^\top 
    \tilde{g}_{t-1}))
\end{equation}

Although we provide the result for vanilla SGD, deriving
the update rule for the case with first order momentum is trivial and 
only adds a multiplicative factor to $\partial \ell/ \partial C$, depending
on the specific implementation of momentum. 
The same analysis for Adam is more involved and most importantly it 
results in the summation in $\partial \nabla \ell(\theta_{t-1})/\partial C_{t-1}$
to lose the appealing property of unitary sensitivity.
Considering also the disparate results of Adam as a DP optimizer 
\cite{mohapatra2022role, andrew2021differentially},
we leave this analysis for future work.
Finally, Algorithm \ref{alg:sota} outlines the online optimization
strategy presented above,
which we call \texttt{OSO-DPSGD}. 

\begin{algorithm}[htbp]
  \begin{algorithmic} [1]
  \caption{Differentially private optimization with \texttt{OSO-DPSGD}}
  \label{alg:sota}
  \State \textbf{Inputs} Samples $z_i \in D$; $\rho$; $T$; $C_0$; $\theta_0$; $|B|$; 
  per-iteration noise multiplier $\nu$; $\nu_q$; $\rho_c$; $\rho_r$; 
  $\tilde{q}_0 = \tilde{g}_0 = 0$.
  \State $\nu_g \leftarrow (\nu^{-2} - \nu_q^{-2})^{-1/2}$
  \For{$t \in \{1, \dots, T\}$}
    \State $B_t \leftarrow$ draw $|B|$ samples uniformly from $D$
    \For{$z_i \in B_t$ \textbf{in parallel}}
        \State $g_t(z_i) \leftarrow \nabla_{\theta_t} \ell(\theta_t, z_i)$
        \State $\bar{g}_t(z_i) \leftarrow g_t(z_i)/\text{max}(1, \frac{\norm{g_t(z_i)}_2}{C_t})$
        \State $q_t(z_i) \leftarrow \frac{g_t(z_i)}{\norm{g_t(z_i)}_2}$ 
         if $\norm{g_t(z_i)}_2>C_t$ else $0$
    \EndFor
    \State $\sigma_g \leftarrow \nu_g C_t$
    \State $\tilde{g}_t \leftarrow \frac{1}{|B|}\left( \sum_{z_i \in B_t} \bar{g}_t(z_i) +
    \mathcal{N}(0, I\sigma_g^2)\right)$
    \State $\theta_{t+1} \leftarrow \theta_t - \rho \hat{g}_t$
    \State $\tilde{q}_t \leftarrow \frac{1}{|B|}\left( \sum_{z_i \in B_t} q_t(z_i) +
    \mathcal{N}(0, I\nu_g^2)\right)$
    \State $C_{t+1} \leftarrow C_t\text{exp}(\rho_c\text{sign}
    (\tilde{g}_{t}{}^\top \tilde{q}_{t-1}))$
    \State $\rho_{t+1} \leftarrow \rho_t\text{exp}(\rho_r\text{sign}(\tilde{g}_{t}{}^\top 
    \tilde{g}_{t-1}))$
  \EndFor
  \end{algorithmic}
\end{algorithm}
In Algorithm \ref{alg:sota} we list the learning rates of $C_t$ and $\rho$ as 
hyperparameters. In practice, especially considering the exponential update rule in 
Equations \eqref{eq:exp_C} and \eqref{eq:exp_r}, they can be set to the same value.
After a qualitative exploration of reasonable values for both, we settle on
$\rho_c = \rho_r = 2.5\cdot 10^{-3}$ for all the experiments.

\section{Experiments} \label{sec:exp}
In the following Section we proceed to assess Algorithm \ref{alg:sota} on a
range of experiments on different datasets, tasks, and model sizes.
In particular, we explore how online sensitivity optimization can be an effective tool
in reducing the privacy and computational costs of running large grid searches. 
In an effort to draw conclusions that can be as general as possible, we identify three
vastly adopted datasets in the literature: MNIST \cite{lecun1998gradient},
FashionMNIST \cite{xiao2017fashion}, and AG News \cite{gulli2005anatomy} 
\cite{zhang2015character}. They are used to train, respectively, a convolutional 
neural network for image classification, a convolutional autoencoder and a bag of words fully
connected neural network for text classification. For further details on the models and their architecture, refer to the Appendix.

\begin{table}[]
\centering
\begin{tabular}{@{}cccc@{}}
\toprule
 & AG News & MNIST & Fashion MNIST \\ \midrule
\begin{tabular}[c]{@{}c@{}}Dataset\\ Size\end{tabular} & 120000 & 60000 & 60000 \\ \midrule
\begin{tabular}[c]{@{}c@{}}Batch\\ Size\end{tabular} & 512 & 512 & 512 \\ \midrule
\begin{tabular}[c]{@{}c@{}}Model \\ Size\end{tabular} & 113156 & 551322 & 48705 \\ \midrule
\end{tabular}
\caption{Dataset and model information shared throughout the experiments.}
\label{tab:1}
\end{table}
Considering the computational burden of benchmarking multiple grid searches, we
devise the following pipeline:
\begin{itemize}
    \item Define the different learning algorithms; to compare \texttt{OSO-DPSGD}
    with relevant strategies, we also include in our experiments the 
    \texttt{FixedThreshold} of \cite{song2013stochastic}
    \cite{shokri2015privacy} \cite{abadi2016deep} among others and 
    \texttt{FixedQuantile} of \cite{andrew2021differentially}. As reported by the 
    respective authors, hyperparameter optimization is performed via grid search over
    the learning rates and threshold values for the former and over the learning rates and
    the quantiles for the latter. Even though \cite{mohapatra2022role} introduce 
    \texttt{AdamWOSM} for the DP adaptive optimization of the learning rate, it still
    tackles the challenge of reducing the number of hyperparameters in a privacy-aware
    grid search, and therefore we include it.
    \item Establish the corresponding grid search ranges. In all of our experiments, we
    fix the ranges of the hyperparameters to the same values. Considering the variety
    of experiments, and without assuming any particular domain knowledge of the task
    at hand, we opt for large ranges: $C \in [10^{-2}, 10^{2}]$ for the clipping
    threshold, $\rho \in [10^{-2.5}, 10^{1.5}]$ for the learning rate and
    $\gamma \in [0.1, 0.9]$ for the target quantile. 
    \item Define grid searches with different granularity. Given the ranges defined in the
    last step, DP training introduces possibly yet another hyperparameter. In fact, 
    increasing the granularity inevitably results in more candidates, and 
    an additional trade off to consider is that of increased fine tuning at the cost of 
    additional privacy leakage. In our experiments, we evaluate 3 grid searches with
    different granularity, i.e. from the $\rho$ and $C$ ranges in the last step we take 
    $k \in \{5, 7, 9\}$ values uniformly separated in a logarithmic scale. For the
    experiments with the \texttt{FixedQuantile} strategy we keep the values
    $\gamma \in [0.1, 0.3, 0.5, 0.7, 0.9]$ defined by the authors in 
    \cite{andrew2021differentially}, as well a setting the learning rate for the
    exponential update rule for $C$ to $0.2$. The initial value for the 
    clipping threshold in both \texttt{FixedQuantile} and \texttt{Online} is set to 
    $C_0 = 0.1$.
    \item Execute private hyperparameter optimization at different privacy levels. 
    For the same $\delta$, we explore with increasing values of $\varepsilon$. 
    Following \cite{mohapatra2022role}, the privacy budgets we establish are per-grid,
    and not per-run. That is, algorithms that need extra fine-tuning and additional 
    parameters, resulting in more runs, will effectively reduce the per-run privacy budget.
    Although this setting may not conform to most past literature, we are motivated
    by approaching DP machine learning from the practitioner point of view, 
    where an oracle providing the optimal hyperparameters may not be a reasonable
    assumption. As in \cite{mohapatra2022role}, we utilize the moment accountant
    to distribute the privacy budget among the configurations, as we do not have a
    large number of candidates.
\end{itemize}

On top of comparing DP learning strategies, we provide a baseline in the non-private
setting, where we iterate only over the learning rate values and initial weights.
To limit the contribution of the Gaussian random noise in the DP setting,
each configuration is executed with $5$ different seeds,
and the results are averaged. Runs with different seeds are not accounted for
in terms of privacy budget.
Given the large number of runs, we validate
each model at training time every $50$ iterations on the full test set, and pick the 
model checkpoint at the best value as representative of the corresponding configuration.
Each configuration runs for $10$ epochs regardless of when the best performance 
is registered. Given the model size and datasets, the total number of epochs is enough to 
have most configurations converge. Nevertheless, we don't expect \textit{every}
combination of hyperparameters to saturate learning, e.g. when training
with $C$ and $\rho$ both set at the lowest value available in the corresponding ranges.
In Tables \ref{tab:mnist}, \ref{tab:fashion}, \ref{tab:agnews}, 
we list the hyperparameters leading to the best results in 
the grid search with granularity $k = 7$ for the corresponding datasets and models. For brevity, we include detailed results only for this specific setting in the main paper.
Results for the remaining granularity values, as well as the numerical results shown graphically only for $k=7$, are deferred to the Appendix.

\begin{figure}[t]
    \centering
    \includegraphics[width=\columnwidth]{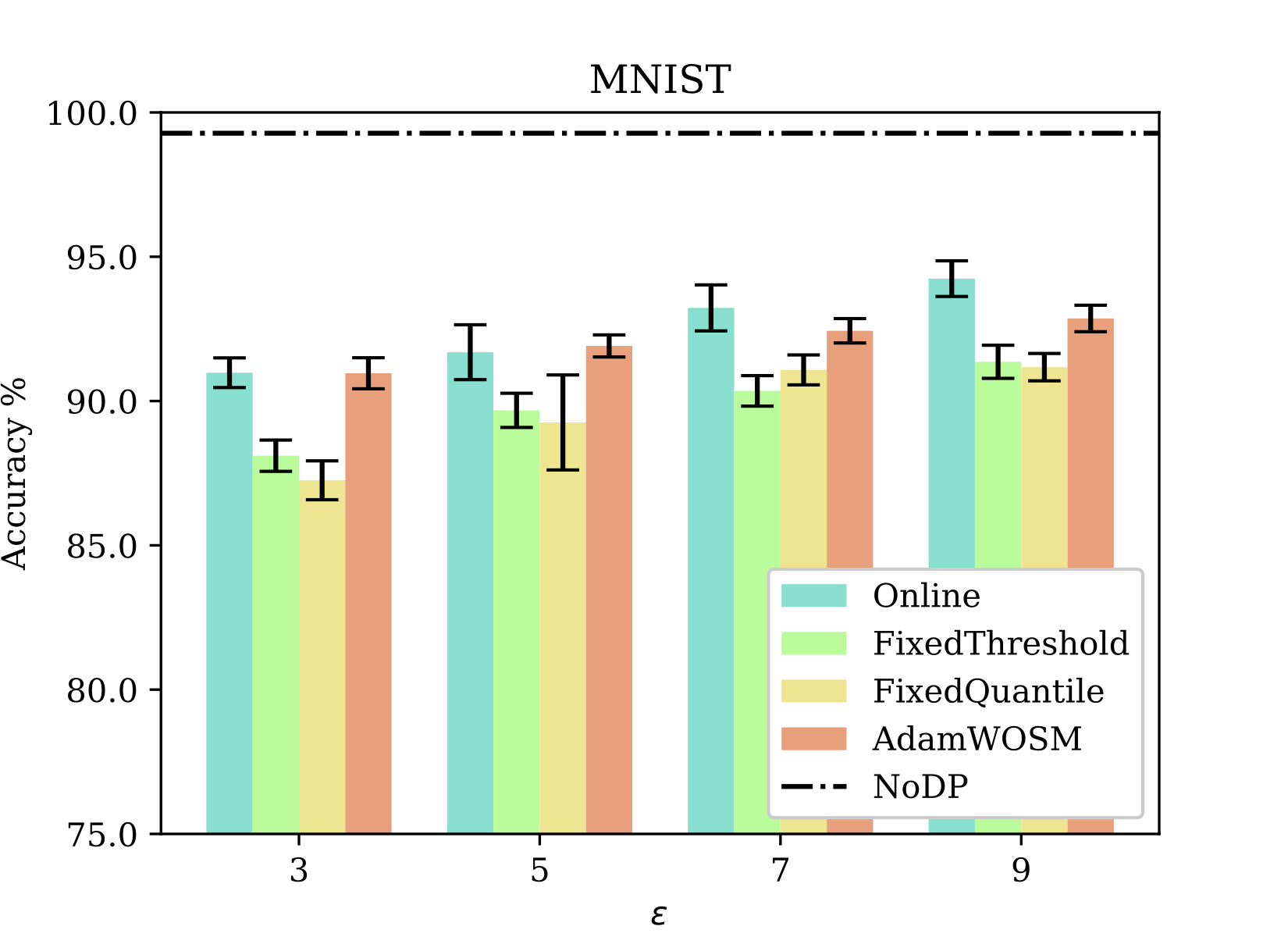}
    \caption{Accuracy on the MNIST dataset. Higher is better.}
    \label{fig1}
\end{figure}

\textbf{Discussion} Figure \ref{fig1} shows the accuracy of the models in the best
configurations, among those tested, on the MNIST dataset. Even though at 
higher privacy levels (low $\varepsilon$) \texttt{Online} and \texttt{AdamWOSM}
appear to be equivalent in terms of results, we can see the former showing better
results when the privacy requirements are relaxed. A possible explanation
may be found in Table \ref{tab:mnist} by noticing that the best $C$ value for
\texttt{AdamWOSM} is fairly large compared to the other strategies. We believe that 
a larger initial value for $C$ may be positive to take long strides towards the direction
of the average gradient at the early stages of the optimization, but may be detrimental
towards the end when reducing the Gaussian noise may help the optimization. 
Nevertheless, we consider both strategies to be roughly equivalent in this experiment.
The results for \texttt{FixedThreshold} and 
\texttt{FixedQuantile} are consistently lower, most likely due to both strategies 
needing a larger grid search, which in turn limits the per-run privacy budget.
Perhaps more surprisingly, the adaptive strategy \texttt{FixedQuantile} does not seem
to show better results compared to fixing the clipping threshold
at the initial value. The improved results that are found in 
\cite{andrew2021differentially} in the federated setting do not seem to translate
in centralized learning, with the experiments we conducted.

\begin{table}[tb]
\centering
\begin{tabular}{@{}cccccccc@{}}
\toprule
 & Online & \multicolumn{2}{c}{\begin{tabular}[c]{@{}c@{}}Fixed\\ Threshold\end{tabular}} & \multicolumn{2}{c}{\begin{tabular}[c]{@{}c@{}}Fixed\\ Quantile\end{tabular}} & \begin{tabular}[c]{@{}c@{}}Adam\\ WOSM\end{tabular} \\ \midrule
$\varepsilon$ & $\rho$ & $\rho$ & $C$ & $\rho^*$ & $\gamma$ & $C$ \\

3 & 0.3162 & 0.01467 & 1.0 & 3.162 & 0.5 & 21.54 \\
5 & 1.467 & 0.003162 & 4.64 & 3.162 & 0.7 & 21.54 \\
7 & 1.467 & 6.812 & 0.010 & 3.162 & 0.7 & 21.54 \\
9 & 1.467 & 6.812 & 0.010 & 3.162 & 0.7 & 21.54 \\ \bottomrule
\end{tabular}
\caption{Best hyperparameters for the MNIST dataset with grid search granularity $k=7$.
Values with ${}^*$ are scaled $\times 10^3$ for better readability. 
Best \texttt{NoDP} result for $\rho = 0.003162$.}
\label{tab:mnist}
\end{table}

\begin{table}[tb]
\begin{tabular}{@{}ccccccc@{}}
\toprule
 & Online & \multicolumn{2}{c}{\begin{tabular}[c]{@{}c@{}}Fixed\\ Threshold\end{tabular}} & \multicolumn{2}{c}{\begin{tabular}[c]{@{}c@{}}Fixed\\ Quantile\end{tabular}} & \begin{tabular}[c]{@{}c@{}}Adam\\ WOSM\end{tabular} \\ \midrule
$\varepsilon$ & $\rho$ & $\rho$ & $C$ & $\rho$ & $\gamma$ & $C$ \\
1 & 0.3162 & 0.0681 & 0.010 & - & - & 0.01 \\
2 & 1.467 & 1.467 & 0.010 & 0.3162 & 0.3 & 0.01 \\
3 & 1.467 & 6.812 & 0.010 & 1.467 & 0.1 & 0.0464 \\
4 & 1.467 & 1.467 & 0.0464 & 1.467 & 0.3 & 0.01 \\ \bottomrule
\end{tabular}
\caption{Best hyperparameters for the Fashion MNIST dataset with grid search granularity $k=7$. 
Best \texttt{NoDP} result for $\rho = 0.01467$. All \texttt{FixedQuantile}
runs diverge for $\varepsilon = 1$.}
\label{tab:fashion}
\end{table}

\begin{table}[tb]
\begin{tabular}{@{}ccccccc@{}}
\toprule
 & Online & \multicolumn{2}{c}{\begin{tabular}[c]{@{}c@{}}Fixed\\ Threshold\end{tabular}} & \multicolumn{2}{c}{\begin{tabular}[c]{@{}c@{}}Fixed\\ Quantile\end{tabular}} & \begin{tabular}[c]{@{}c@{}}Adam\\ WOSM\end{tabular} \\ \midrule
$\varepsilon$ & $\rho$ & $\rho$ & $C$ & $\rho*$ & $\gamma$ & $C$ \\
3 & 0.06812 & 1.467 & 0.01 & 3.162 & 0.5 & 0.01 \\
5 & 0.06812 & 1.467 & 0.010 & 3.162 & 0.5 & 0.01 \\
7 & 0.06812 & 1.467 & 0.010 & 3.162 & 0.7 & 0.01 \\
9 & 0.06812 & 0.03162 & 0.0464 & 3.162 & 0.7 & 0.01 \\ \bottomrule
\end{tabular}
\caption{Best hyperparameters for the AG News dataset.
Values with ${}^*$ are scaled $\times 10^3$ for better readability.
$k=7$. Best \texttt{NoDP} result for $\rho = 0.003162$.}
\label{tab:agnews}
\end{table}

\begin{figure}[t]
    \centering
    \includegraphics[width=\columnwidth]{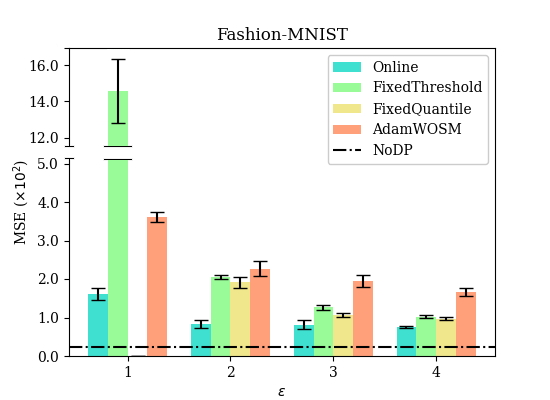}
    \caption{Mean Squared Error on the Fashion MNIST dataset. Lower is better.
    All runs for $\varepsilon=1$ of \texttt{FixedQuantile} result in a diverging
    optimization and are therefore not included.}
    \label{fig2}    
\end{figure}

Figure \ref{fig2} shows the best results in terms of mean squared error on the 
FashionMNIST dataset, where a model is trained to encode and decode the
input images of clothing items. The chosen architecture is based on a
convolutional autoencoder, and it has the smallest number of parameters among
those considered in this work, as in Table \ref{tab:1}.
The privacy regimes are then chosen accordingly. Firstly, we notice that
for $\varepsilon=1$ the \texttt{FixedQuantile} strategy does not converge
with any of the available hyperparameters. To justify this result, we highlight
how in Table \ref{tab:fashion} all other strategies adopt aggressive clipping 
strategies with small $C$'s. We thus believe that for very high privacy regimes
even running with $\gamma=0.1$ (the lowest value for the target quantile) may 
induce large swings in the exponential updates of $C_t$, disrupting the
optimization. Nevertheless, for $\varepsilon \in \{2, 3, 4\}$ this strategy shows
the second best results. Conversely, \texttt{AdamWOSM} may be penalized by the 
choice of the initial $\rho_0 = 10^{-3}$, as suggested by the authors in 
\cite{mohapatra2022role}. In fact, we notice from Table \ref{tab:agnews} that 
the optimal clipping threshold is very small in all competing strategies,
and the combination of small $C$ and small $\rho_0$ may render the 
optimization excessively slow to converge within the set number of epochs. 
Further, it may suggest that adapting the learning rate on a per-parameter basis,
as in \texttt{AdamWOSM}, can be effective as long as the base learning 
rate is itself carefully selected. Thus, optimizing $\rho_t$ in the grid
search, and then adaptively tuning it within the same run, as done in  
\texttt{Online}, seems to show better results.

\begin{figure}[t]
    \centering
    \includegraphics[width=\columnwidth]{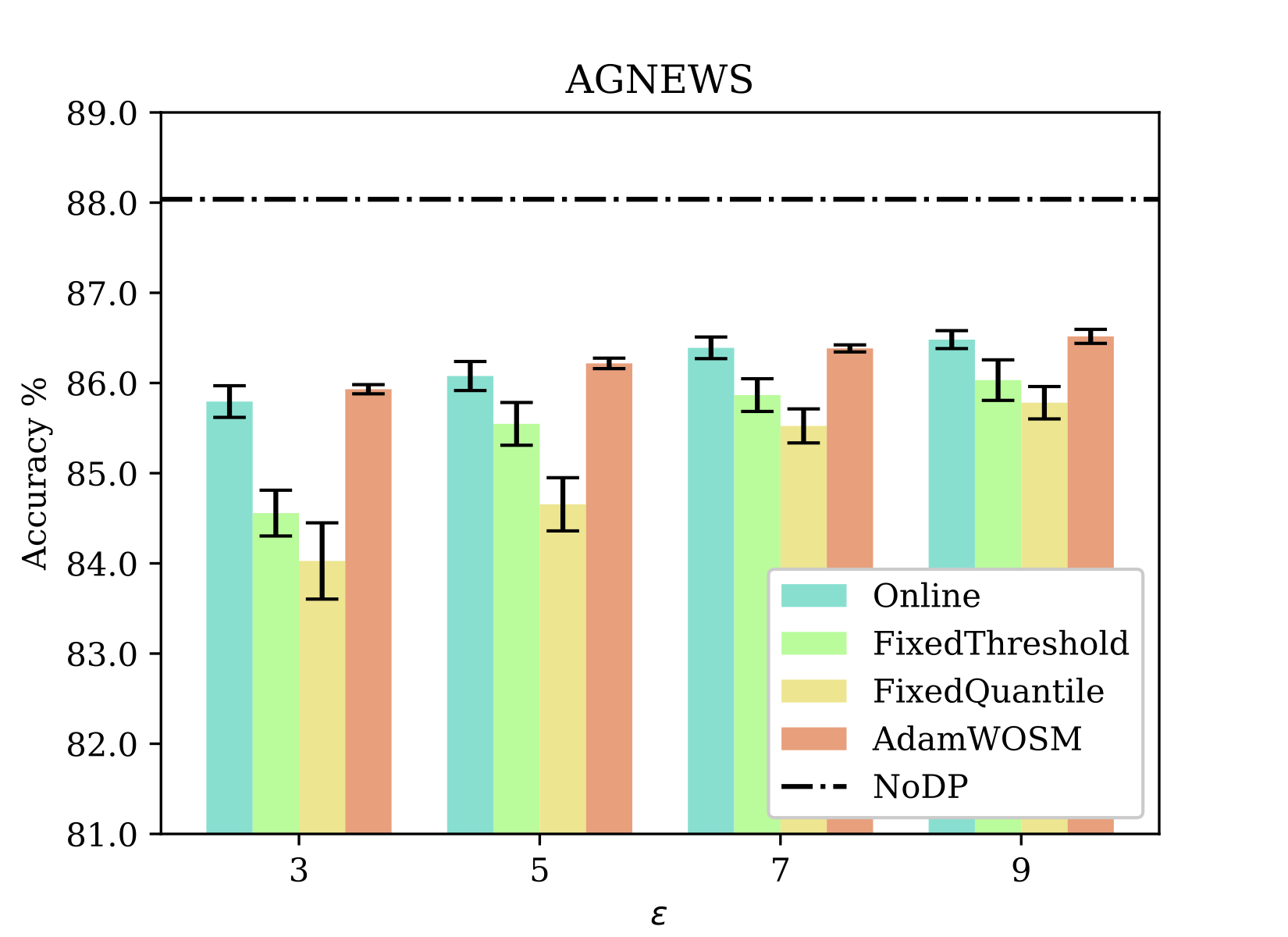}
    \caption{Accuracy on the AG News dataset. Higher is better.}
    \label{fig3}
\end{figure}

Figure \ref{fig3} plots the accuracy on the AG News dataset, where a bag of 
words model with a fully connected neural network is used to classify a 
selection of news in one of four classes. In this experiment we notice that 
\texttt{AdamWOSM} performs the best, with \texttt{Online} being marginally
below. Still, as with the MNIST dataset, we take both strategies 
to be comparable in these two settings, as the average of one roughly fits within a 
standard deviation of the other.


\section{Conclusion} \label{sec:conc}
This work studies differentially private machine learning in the context of
hyperparameter optimization, where the privacy cost of running a grid search 
is accounted for. Under these conditions, algorithms that require one less parameter
may be preferable. Thus we explore strategies for the adaptive tuning of the clipping
threshold $C$, and derive a result inspired by online learning rate optimization. With 
the proposed strategy, which we incorporate in the \texttt{OSO-DPSGD} algorithm, the
clipping threshold is updated at each iteration based on the direction of steepest descent
of the cost function. The resulting update rule is particularly clean, and results 
in the dot product between two sanitized vector queries:
the average gradient at time $t$, and the derivative
w.r.t. $C$ of the gradient at time $t-1$. With the former already needed in standard 
\texttt{DP-SGD}, and the latter resulting in a query with unitary sensitivity, the
additional computational and privacy burden is minimal.
Our range of experiments seems to encourage further research in this area, as 
online sensitivity optimization shows comparable results with one less
parameter when assessed against standard state of the art algorithms,
if the privacy guarantees are required at a grid search level, and not just within
a single run. In the future, we hope to refine our analysis and algorithm, to possibly achieve
better results even in this latter setting of per-run privacy requirements.

\section{Acknowledgments}
The work of Catuscia Palamidessi was supported by the European Research Council (ERC) grant Hypatia 
(grant agreement N. 835294) under the European Union’s Horizon 2020 research and innovation programme.

\bibliography{sample}

\appendix
\clearpage

\section{Appendix - Models and Experiments}

\begin{figure}[tbh]
    \centering
    \includegraphics[width=\columnwidth]{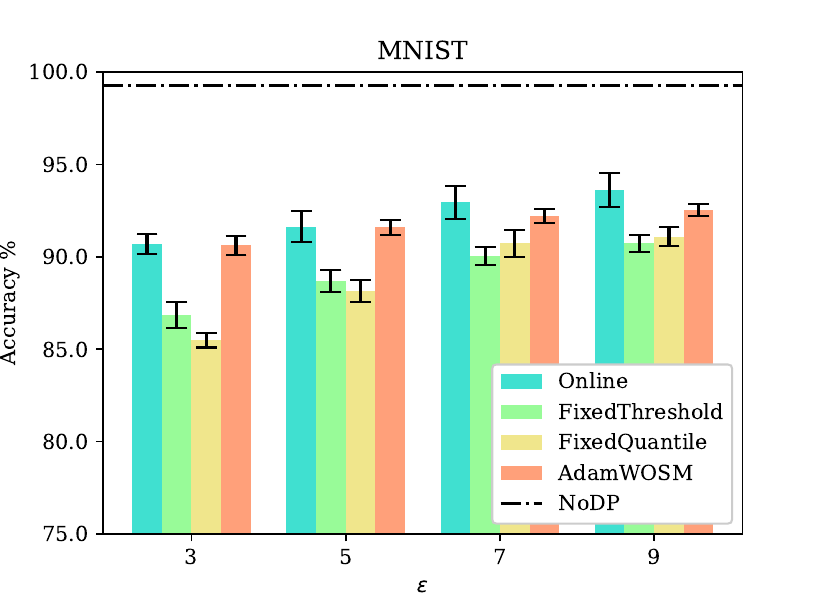}
    \caption{$k = 9$. Accuracy on the MNIST dataset. Higher is better.
    Refer to Table \ref{tab:k9-mnist} for numeric results and optimized
    hyperparameters.}
    \label{fig:k9-mnist}
\end{figure}

\begin{figure}[tbh]
    \centering
    \includegraphics[width=\columnwidth]{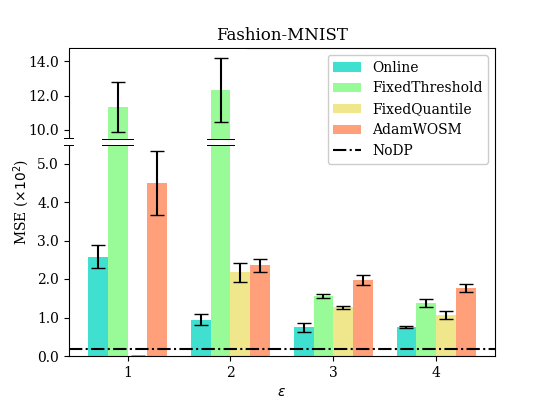}
    \caption{$k = 9$. Mean Squared Error on the Fashion MNIST dataset. Lower is better. Refer to Table \ref{tab:k9-fashion} for numeric results and optimized
    hyperparameters.}
    \label{fig:k9-fashion}
\end{figure}

\begin{figure}[tbh]
    \centering
    \includegraphics[width=\columnwidth]{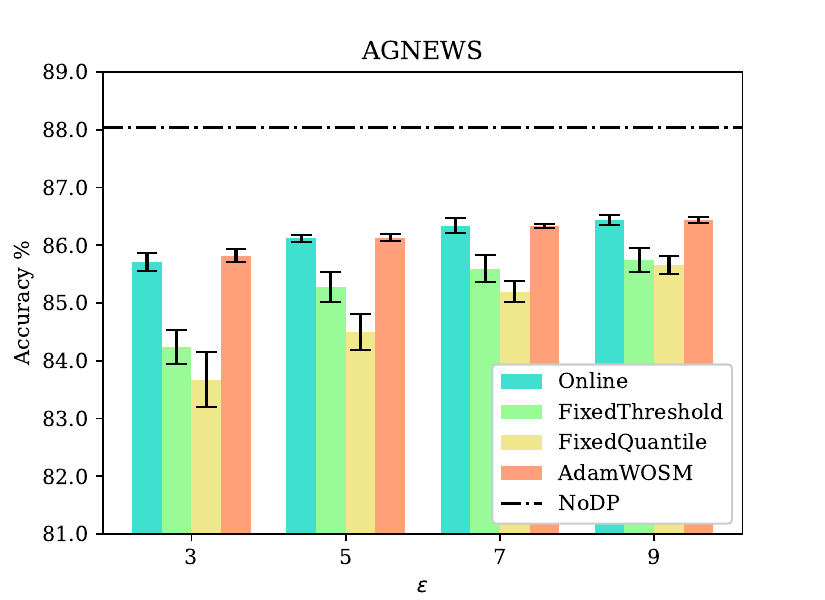}
    \caption{$k = 9$. Accuracy on the AG News dataset. Higher is better. 
    Refer to Table \ref{tab:k9-agnews} for numeric results and optimized
    hyperparameters.}
    \label{fig:k9-agnews}
\end{figure}

\begin{figure}[tbh]
    \centering
    \includegraphics[width=\columnwidth]{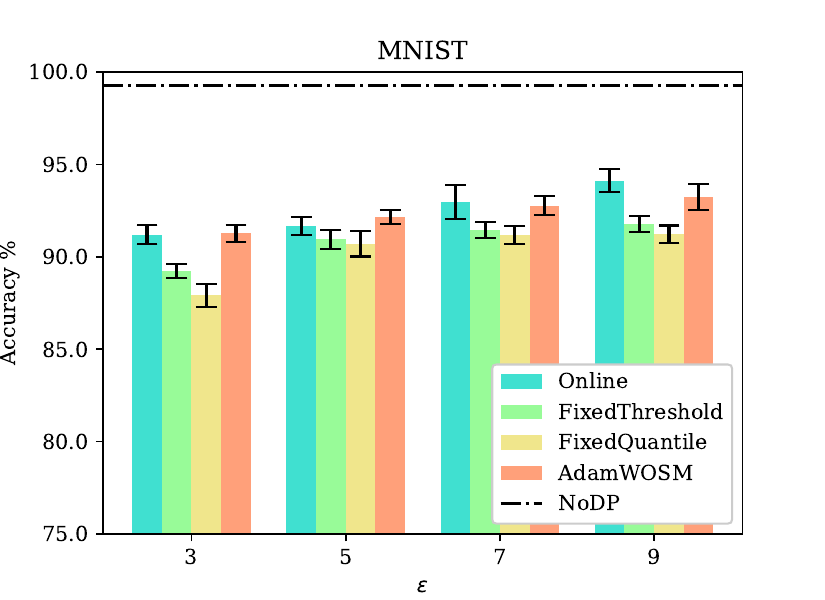}
    \caption{$k = 5$. Accuracy on the MNIST dataset. Higher is better.
    Refer to Table \ref{tab:k5-mnist} for numeric results and optimized
    hyperparameters.}
    \label{fig:k5-mnist}
\end{figure}

\begin{figure}[tbh]
    \centering
    \includegraphics[width=\columnwidth]{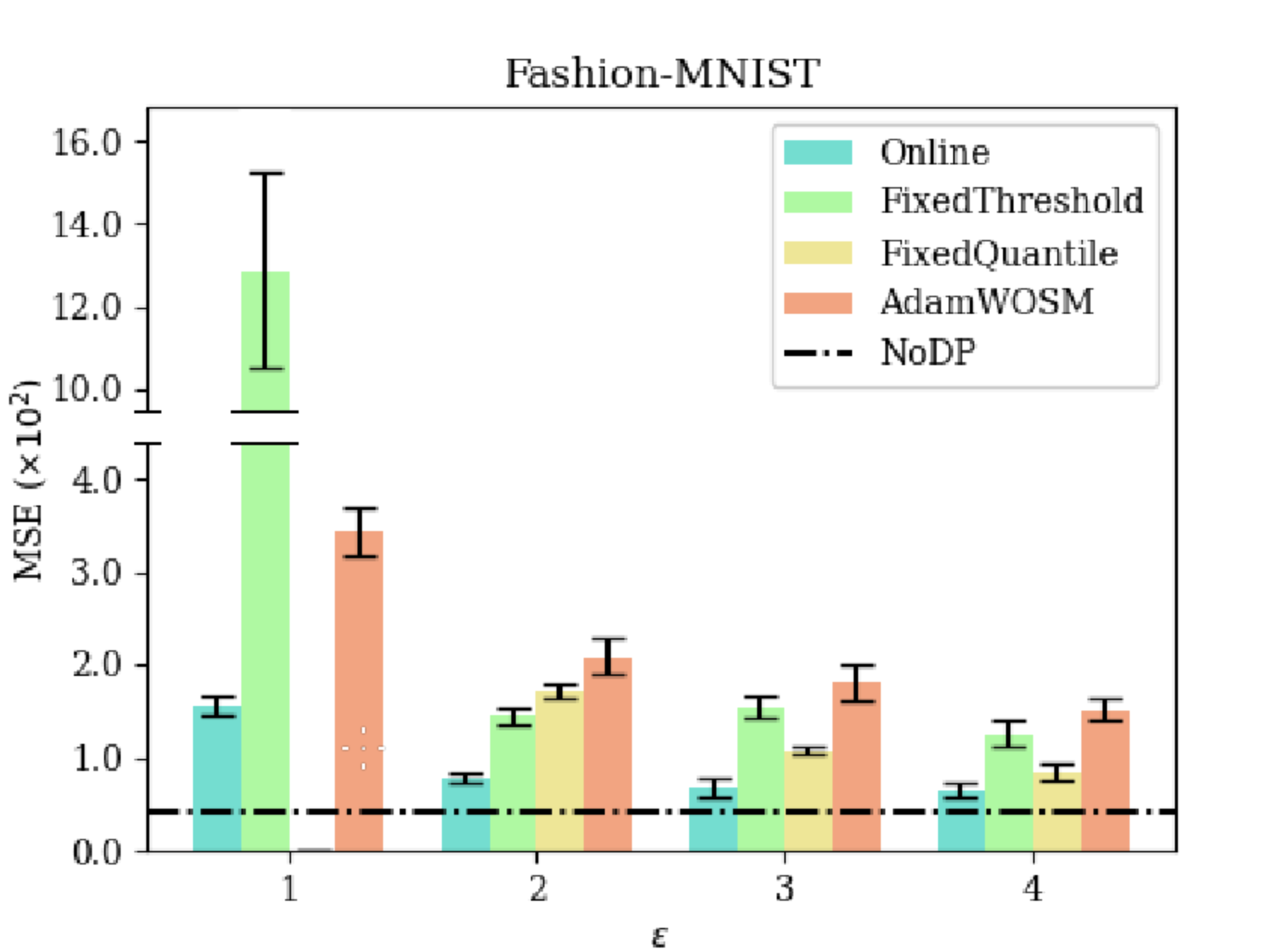}
    \caption{$k = 5$. Mean Squared Error on the Fashion MNIST dataset. Lower is better.
    Refer to Table \ref{tab:k5-fashion} for numeric results and optimized
    hyperparameters.}
    \label{fig:k5-fashion}
\end{figure}

\begin{figure}[tbh]
    \centering
    \includegraphics[width=\columnwidth]{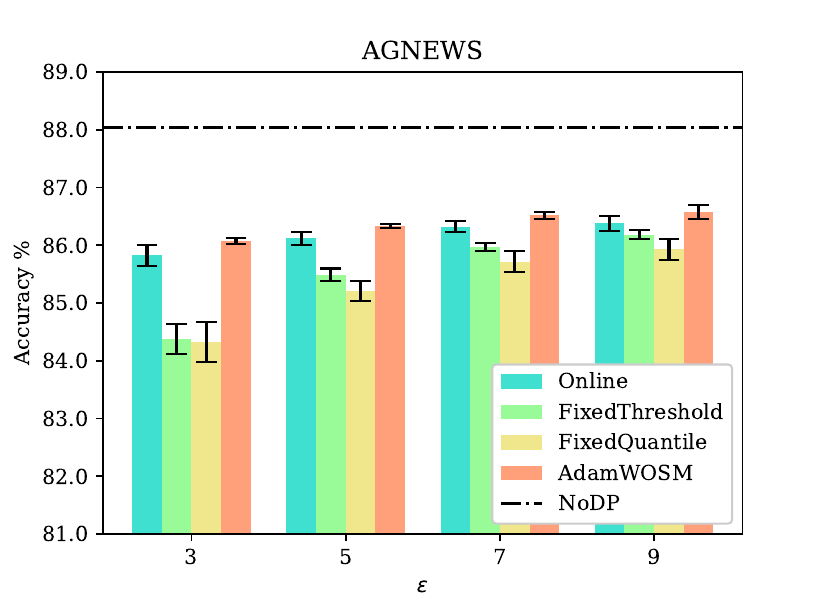}
    \caption{$k = 5$. Accuracy on the AG News dataset. Higher is better.
    Refer to Table \ref{tab:k5-agnews} for numeric results and optimized
    hyperparameters.}
    \label{fig:k5-agnews}
\end{figure}

As mentioned in the corresponding Section of the main paper, we provide  
additional details about the model architectures, datasets, and results of 
experiments at different granularity levels.
The model architectures are outlined in Tables \ref{tab:cnn}, \ref{tab:ae}, \ref{tab:bow}.
All datasets go through minor pre-processing, that is pixel values are mapped to the $[0, 1]$
interval, while the text-based dataset AG News first goes through word embedding, using
an embedding size of $50$ for up to the first $25$ words. To speed up development, we
use the pre-trained word embeddings from \cite{pennington2014glove}.
Next, we report the results for granularity $k = 5$
in Tables \ref{tab:k5-mnist}, \ref{tab:k5-fashion} \ref{tab:k5-agnews} and 
Figures \ref{fig:k5-mnist}, \ref{fig:k5-fashion} and \ref{fig:k5-agnews}.
For $k = 9$, results are presented in Tables \ref{tab:k9-mnist}, 
\ref{tab:k9-fashion}, \ref{tab:k9-agnews} and 
Figures \ref{fig:k9-mnist}, \ref{fig:k9-fashion} and \ref{fig:k9-agnews}.

\begin{table}[tbh]
\centering
\begin{tabular}{@{}cccc@{}}
\toprule
\multicolumn{4}{c}{CNN} \\ \midrule
layer & \begin{tabular}[c]{@{}c@{}}kernel size\\ output size\end{tabular} & stride & \begin{tabular}[c]{@{}c@{}}non\\ linearity\end{tabular} \\
\midrule
2D Conv & $16\times8\times8$ & $1\times1$ & ReLU \\
2D MaxPool & $2\times2$ & $1\times1$ & - \\
2D Conv & $32\times4\times4$ & $1\times1$ & ReLU \\
Linear & $32$ & - & ReLU \\
Linear & $10$ & - & ReLU \\
\bottomrule
\end{tabular}
\caption{CNN}
\label{tab:cnn}
\end{table}

\begin{table}[tbh]
\begin{tabular}{@{}cccc@{}}
\toprule
\multicolumn{4}{c}{AutoEncoder} \\ \midrule
layer & \begin{tabular}[c]{@{}c@{}}kernel size\\ output size\end{tabular} & stride & \begin{tabular}[c]{@{}c@{}}non\\ linearity\end{tabular} \\ \midrule
2D Conv & $8\times3\times3$ & $1\times1$ & LeakyReLU \\
2D Conv & $16\times3\times3$ & $1\times1$ & LeakyReLU \\
2D Conv & $32\times3\times3$ & $1\times1$ & LeakyReLU \\
2D Conv & $64\times3\times3$ & $1\times1$ & LeakyReLU \\
2D Transpose Conv & $32\times3\times3$ & $1\times1$ & LeakyReLU \\
2D Transpose Conv & $16\times3\times3$ & $1\times1$ & LeakyReLU \\
2D Transpose Conv & $8\times3\times3$ & $1\times1$ & LeakyReLU \\
2D Transpose Conv & $1\times3\times3$ & $1\times1$ & Sigmoid \\ \bottomrule
\end{tabular}
\caption{AutoEncoder}
\label{tab:ae}
\end{table}

\begin{table}[]
\centering
\begin{tabular}{@{}ccc@{}}
\toprule
\multicolumn{3}{c}{BagOfWords - FC} \\ \midrule
layer & output size & \begin{tabular}[c]{@{}c@{}}non\\ linearity\end{tabular} \\ \midrule
Linear & $128$ & LeakyReLU \\
Linear & $128$ & LeakyReLU \\
Linear & $4$ & LeakyReLU  \\
\bottomrule
\end{tabular}
\caption{Bag of Words model architecture with a fully connected neural network.}
\label{tab:bow}
\end{table}


\begin{table*}[tb]
\begin{tabular}{@{}c|ccc|cccc|cccc|ccc@{}}
\toprule
 & \multicolumn{3}{c|}{Online} & \multicolumn{4}{c|}{\begin{tabular}[c]{@{}c@{}}Fixed\\ Threshold\end{tabular}} & \multicolumn{4}{c|}{\begin{tabular}[c]{@{}c@{}}Fixed\\ Quantile\end{tabular}} & \multicolumn{3}{c}{\begin{tabular}[c]{@{}c@{}}Adam\\ WOSM\end{tabular}} \\ \midrule
$\varepsilon$ & $\rho$ & acc & std dev & $\rho$ & $C$ & acc & std dev & $\rho^*$ & $\gamma$ & acc & std dev & $C$ & acc & std dev \\
3 & 0.316 & 90.69 & 0.53 & 100.0 & 0.1000 & 86.83 & 0.70 & 3.162 & 0.5 & 85.48 & 0.39 & 10.0 & 90.62 & 0.52 \\
5 & 1.00 & 91.62 & 0.83 & 3.162 & 3.162 & 88.69 & 0.57 & 3.162 & 0.5 & 88.13 & 0.59 & 10.0 & 91.60 & 0.39 \\
7 & 1.00 & 92.94 & 0.90 & 3162.0 & 0.0100 & 90.04 & 0.49 & 3.162 & 0.7 & 90.73 & 0.72 & 31.62 & 92.19 & 0.37 \\
9 & 1.00 & 93.62 & 0.91 & 3.162 & 10.00 & 90.72 & 0.47 & 3.162 & 0.7 & 91.09 & 0.51 & 31.62 & 92.53 & 0.33 \\ \bottomrule
\end{tabular}
\caption{MNIST, $k=9$, best \texttt{NoDP} for $\rho = 0.003162$, ${}^*$ values 
are scaled $\times 10^3$.}
\label{tab:k9-mnist}
\end{table*}

\begin{table*}[]
\begin{tabular}{@{}c|ccc|cccc|cccc|ccc@{}}
\toprule
 & \multicolumn{3}{c|}{Online} & \multicolumn{4}{c|}{\begin{tabular}[c]{@{}c@{}}Fixed\\ Threshold\end{tabular}} & \multicolumn{4}{c|}{\begin{tabular}[c]{@{}c@{}}Fixed\\ Quantile\end{tabular}} & \multicolumn{3}{c}{\begin{tabular}[c]{@{}c@{}}Adam\\ WOSM\end{tabular}} \\ \midrule
$\varepsilon$ & $\rho$ & \begin{tabular}[c]{@{}c@{}}mse\\ $\times 10^2$\end{tabular} & std dev & $\rho$ & $C$ & \begin{tabular}[c]{@{}c@{}}mse\\ $\times 10^2$\end{tabular} & std dev & $\rho$ & $\gamma$ & \begin{tabular}[c]{@{}c@{}}mse\\ $\times 10^2$\end{tabular} & std dev & $C$ & \begin{tabular}[c]{@{}c@{}}mse\\ $\times 10^2$\end{tabular} & std dev \\
1 & 0.100 & 2.57 & 0.29 & 0.0316 & 0.01 & 11.32 & 1.46 & - & - & - & - & 0.0100 & 4.48 & 0.83 \\
2 & 1.000 & 0.94 & 0.13 & 0.100 & 0.01 & 12.35 & 1.87 & 0.316 & 0.3 & 2.17 & 0.25 & 0.0100 & 2.35 & 0.16 \\
3 & 3.162 & 0.74 & 0.10 & 0.3162 & 0.10 & 1.54 & 0.04 & 1.000 & 0.1 & 1.26 & 0.04 & 0.0316 & 1.97 & 0.13 \\
4 & 3.162 & 0.76 & 0.02 & 1.000 & 0.10 & 1.37 & 0.10 & 1.000 & 0.3 & 1.05 & 0.10 & 0.0316 & 1.76 & 0.10 \\ \bottomrule
\end{tabular}
\caption{FashionMNIST, $k=9$, best \texttt{NoDP} for $\rho = 0.003162$.}
\label{tab:k9-fashion}
\end{table*}

\begin{table*}[]
\begin{tabular}{@{}c|ccc|cccc|cccc|ccc@{}}
\toprule
 & \multicolumn{3}{c|}{Online} & \multicolumn{4}{c|}{\begin{tabular}[c]{@{}c@{}}Fixed\\ Threshold\end{tabular}} & \multicolumn{4}{c|}{\begin{tabular}[c]{@{}c@{}}Fixed\\ Quantile\end{tabular}} & \multicolumn{3}{c}{\begin{tabular}[c]{@{}c@{}}Adam\\ WOSM\end{tabular}} \\ \midrule
$\varepsilon$ & $\rho$ & acc & std dev & $\rho$ & $C$ & acc & std dev & $\rho^*$ & $\gamma$ & acc & std dev & $C$ & acc & std dev \\
3 & 0.1 & 85.70 & 0.15 & 1.000 & 0.0100 & 84.23 & 0.29 & 3.162 & 0.5 & 83.67 & 0.48 & 0.0100 & 85.82 & 0.11 \\
5 & 0.1 & 86.12 & 0.05 & 0.316 & 0.0316 & 85.27 & 0.25 & 3.162 & 0.5 & 84.49 & 0.31 & 0.0100 & 86.13 & 0.05 \\
7 & 0.1 & 86.34 & 0.12 & 0.100 & 0.1000 & 85.59 & 0.23 & 3.162 & 0.7 & 85.19 & 0.17 & 0.0100 & 86.33 & 0.03 \\
9 & 0.1 & 86.43 & 0.08 & 0.316 & 0.0316 & 85.74 & 0.20 & 3.162 & 0.7 & 85.65 & 0.15 & 0.0316 & 86.43 & 0.05 \\ \bottomrule
\end{tabular}
\caption{AG News, $k=9$, ${}^*$ values are scaled $\times 10^3$, best \texttt{NoDP} for $\rho = 0.003162$.}
\label{tab:k9-agnews}
\end{table*}

\begin{table*}[]
\begin{tabular}{@{}c|ccc|cccc|cccc|ccc@{}}
\toprule
 & \multicolumn{3}{c|}{Online} & \multicolumn{4}{c|}{\begin{tabular}[c]{@{}c@{}}Fixed\\ Threshold\end{tabular}} & \multicolumn{4}{c|}{\begin{tabular}[c]{@{}c@{}}Fixed\\ Quantile\end{tabular}} & \multicolumn{3}{c}{\begin{tabular}[c]{@{}c@{}}Adam\\ WOSM\end{tabular}} \\ \midrule
$\varepsilon$ & $\rho$ & acc & std dev & $\rho*$ & $C$ & acc & std dev & $\rho^*$ & $\gamma$ & acc & std dev & $C$ & acc & std dev \\
3 & 0.316 & 91.19 & 0.51 & 3162.0 & 0.01 & 89.2 & 0.39 & 3.162 & 0.5 & 87.91 & 0.62 & 10.0 & 91.27 & 0.46 \\
5 & 0.316 & 91.66 & 0.50 & 3.162 & 10.00 & 90.9 & 0.49 & 3.162 & 0.7 & 90.71 & 0.69 & 10.0 & 92.16 & 0.38 \\
7 & 3.162 & 92.96 & 0.90 & 3.162 & 10.00 & 91.4 & 0.43 & 3.162 & 0.7 & 91.16 & 0.49 & 10.0 & 92.76 & 0.52 \\
9 & 3.162 & 94.11 & 0.63 & 3.162 & 10.00 & 91.7 & 0.42 & 3.162 & 0.7 & 91.21 & 0.47 & 10.0 & 93.22 & 0.70 \\ \bottomrule
\end{tabular}
\caption{MNIST, $k=5$, ${}^*$ values are scaled $\times 10^3$, best \texttt{NoDP} for $\rho = 0.003162$.}
\label{tab:k5-mnist}
\end{table*}

\begin{table*}[]
\begin{tabular}{@{}c|ccc|cccc|cccc|ccc@{}}
\toprule
 & \multicolumn{3}{c|}{Online} & \multicolumn{4}{c|}{\begin{tabular}[c]{@{}c@{}}Fixed\\ Threshold\end{tabular}} & \multicolumn{4}{c|}{\begin{tabular}[c]{@{}c@{}}Fixed\\ Quantile\end{tabular}} & \multicolumn{3}{c}{\begin{tabular}[c]{@{}c@{}}Adam\\ WOSM\end{tabular}} \\ \midrule
$\varepsilon$ & $\rho$ & \begin{tabular}[c]{@{}c@{}}mse\\ $\times 10^2$\end{tabular} & std dev & $\rho$ & $C$ & \begin{tabular}[c]{@{}c@{}}mse\\ $\times 10^2$\end{tabular} & std dev & $\rho$ & $\gamma$ & \begin{tabular}[c]{@{}c@{}}mse\\ $\times 10^2$\end{tabular} & std dev & $C$ & \begin{tabular}[c]{@{}c@{}}mse\\ $\times 10^2$\end{tabular} & std dev \\
3 & 0.316 & 1.56 & 0.10 & 0.0316 & 0.01 & 12.88 & 2.37 & - & - & - & - & 0.01 & 3.43 & 0.26 \\
5 & 3.162 & 0.77 & 0.05 & 3.162 & 0.01 & 1.44 & 0.08 & 0.316 & 0.3 & 1.71 & 0.08 & 0.01 & 2.09 & 0.18 \\
7 & 3.162 & 0.67 & 0.09 & 0.3162 & 0.10 & 1.54 & 0.10 & 3.162 & 0.1 & 1.06 & 0.03 & 0.01 & 1.81 & 0.18 \\
9 & 3.162 & 0.64 & 0.07 & 31.62 & 0.01 & 1.25 & 0.14 & 3.162 & 0.1 & 0.84 & 0.08 & 0.01 & 1.51 & 0.11 \\ \bottomrule
\end{tabular}
\caption{Fashion MNIST, $k=5$, best \texttt{NoDP} for $\rho = 0.003162$.}
\label{tab:k5-fashion}
\end{table*}

\begin{table*}[]
\begin{tabular}{@{}c|ccc|cccc|cccc|ccc@{}}
\toprule
 & \multicolumn{3}{c|}{Online} & \multicolumn{4}{c|}{\begin{tabular}[c]{@{}c@{}}Fixed\\ Threshold\end{tabular}} & \multicolumn{4}{c|}{\begin{tabular}[c]{@{}c@{}}Fixed\\ Quantile\end{tabular}} & \multicolumn{3}{c}{\begin{tabular}[c]{@{}c@{}}Adam\\ WOSM\end{tabular}} \\ \midrule
$\varepsilon$ & $\rho$ & acc & std dev & $\rho$ & $C$ & acc & std dev & $\rho$ & $\gamma$ & acc & std dev & $C$ & acc & std dev \\
3 & 0.0316 & 85.82 & 0.18 & 3.162 & 0.01 & 84.37 & 0.25 & 3.162 & 0.5 & 84.32 & 0.34 & 0.01 & 86.07 & 0.05 \\
5 & 0.3162 & 86.12 & 0.11 & 3.162 & 0.01 & 85.49 & 0.10 & 3.162 & 0.7 & 85.20 & 0.17 & 0.01 & 86.33 & 0.03 \\
7 & 0.3162 & 86.32 & 0.09 & 3.162 & 0.01 & 85.97 & 0.07 & 3.162 & 0.7 & 85.71 & 0.18 & 0.01 & 86.51 & 0.06 \\
9 & 0.3162 & 86.38 & 0.12 & 3.162 & 0.01 & 86.18 & 0.07 & 3.162 & 0.7 & 85.93 & 0.18 & 0.01 & 86.57 & 0.11 \\ \bottomrule
\end{tabular}
\caption{AG News, $k=5$, best \texttt{NoDP} for $\rho = 0.003162$.}
\label{tab:k5-agnews}
\end{table*}

\end{document}